\pdfoutput=1

\documentclass[11pt]{article}

\usepackage[]{ACL2023}
\usepackage{times}
\usepackage{latexsym}

\usepackage[T1]{fontenc}

\usepackage[utf8]{inputenc}

\usepackage{microtype}

\usepackage{inconsolata}

\usepackage{amsthm}
\usepackage{amsfonts}
\usepackage{amssymb}
\usepackage{mathtools}
\usepackage{amscd}
\usepackage{bm}
\usepackage{enumitem}
\usepackage{booktabs}
\usepackage{geometry}
\usepackage{subcaption}
\usepackage{multirow}
\usepackage{geometry}
\usepackage{pifont}
\usepackage{subcaption}
\usepackage{afterpage}

\newtheorem{theorem}{Theorem}[section]

\newtheorem{lemma}[theorem]{Lemma}
\newtheorem{proposition}[theorem]{Proposition}
\usepackage{graphicx}
\usepackage{float}
\usepackage{algorithm}
\usepackage{algpseudocode}
\newfloat{method}{htbp}{lop}
\floatname{method}{Method}

\usepackage[usenames, dvipsnames]{xcolor} 
\title{API Is Enough: Conformal Prediction for Large Language Models Without Logit-Access}

\author{Jiayuan Su\textsuperscript{1}, Jing Luo\textsuperscript{2}, Hongwei Wang\textsuperscript{1}$^{*}$, Lu Cheng\textsuperscript{2}\thanks{*Corresponding authors.} \\
  \textsuperscript{1}Zhejiang University - University of Illinois Urbana-Champaign Institute, Zhejiang University\\
  \textsuperscript{2}Department of Computer Science, University of Illinois Chicago 
  \\
  \texttt{\{jiayuan.23, hongweiwang\}@intl.zju.edu.cn,} \\
  \texttt{\{jluo31, lucheng\}@uic.edu}
}

\begin{document}
\maketitle

\begin{abstract}
This study aims to address the pervasive challenge of quantifying uncertainty in large language models (LLMs) without logit-access. Conformal Prediction (CP), known for its model-agnostic and distribution-free features, is a desired approach for various LLMs and data distributions. However, existing CP methods for LLMs typically assume access to the logits, which are unavailable for some API-only LLMs. In addition, logits are known to be miscalibrated, potentially leading to degraded CP performance. To tackle these challenges, we introduce a novel CP method that (1) is tailored for API-only LLMs without logit-access; (2) minimizes the size of prediction sets; and (3) ensures a statistical guarantee of the user-defined coverage. The core idea of this approach is to formulate nonconformity measures using both coarse-grained (i.e., sample frequency) and fine-grained uncertainty notions (e.g., semantic similarity). Experimental results on both close-ended and open-ended Question Answering tasks show our approach can mostly outperform the logit-based CP baselines.
\end{abstract}
\section{Introduction}
Large Language Models (LLMs) have made significant advancements \cite{thoppilan2022lamda, wei2022chain, wei2023larger}, highlighting the research potential of natural language generation \cite{peinl2023evaluation}. However, they often generate information that is not accurate, factual, or grounded in reality, referred to as "hallucination" \cite{lecun2023large}. Therefore, it is crucial to quantify LLM uncertainty to ensure responsible responses.

However, uncertainty quantification (UQ) for LLMs is challenging due to the complex data distributions and inner model mechanism, as well as the often limited access to logit information. A potential solution is to use conformal prediction (CP) \cite{vovk2005algorithmic, angelopoulos2021gentle, kato2023review,wang2023equal}, which is known for being model-agnostic and distribution-free, and with rigorous coverage guarantees. Given a user-defined error rate $\alpha$, CP provides a guaranteed coverage rate for prediction sets/intervals. It measures the uncertainty from a model prediction using nonconformity score functions, e.g., $1 - f(X)_Y$ \cite{sadinle2019least}, where $f(X)_Y$ is the softmax score for the true label $Y$. 

Most of the existing CPs for LLMs rely on the access to model logits to measure nonconformity scores. For instance, \citet{kumar2023conformal} define nonconformity scores as softmax scores for logits of different options in the multi-choice question answering (MCQ) task and \citet{quach2023conformal} apply the conformal risk control framework \cite{angelopoulos2021learn}, an extension of CP, to LLMs by utilizing model-based log probability. However, for some API-only LLMs like Bard \cite{manyika2023overview}, logit-access is almost impossible for end users. Even though the logits are available (e.g., for GPT 4V \cite{openai2023gpt4v}), they are known to be miscalibrated and can lead to degraded performance of CP w.r.t. estimating the prediction sets or intervals \cite{nguyen2015posterior, lin2022teaching}, e.g., a large set size (i.e., low efficiency).


\begin{figure*}[t]
\centering
    \includegraphics[width=\linewidth]{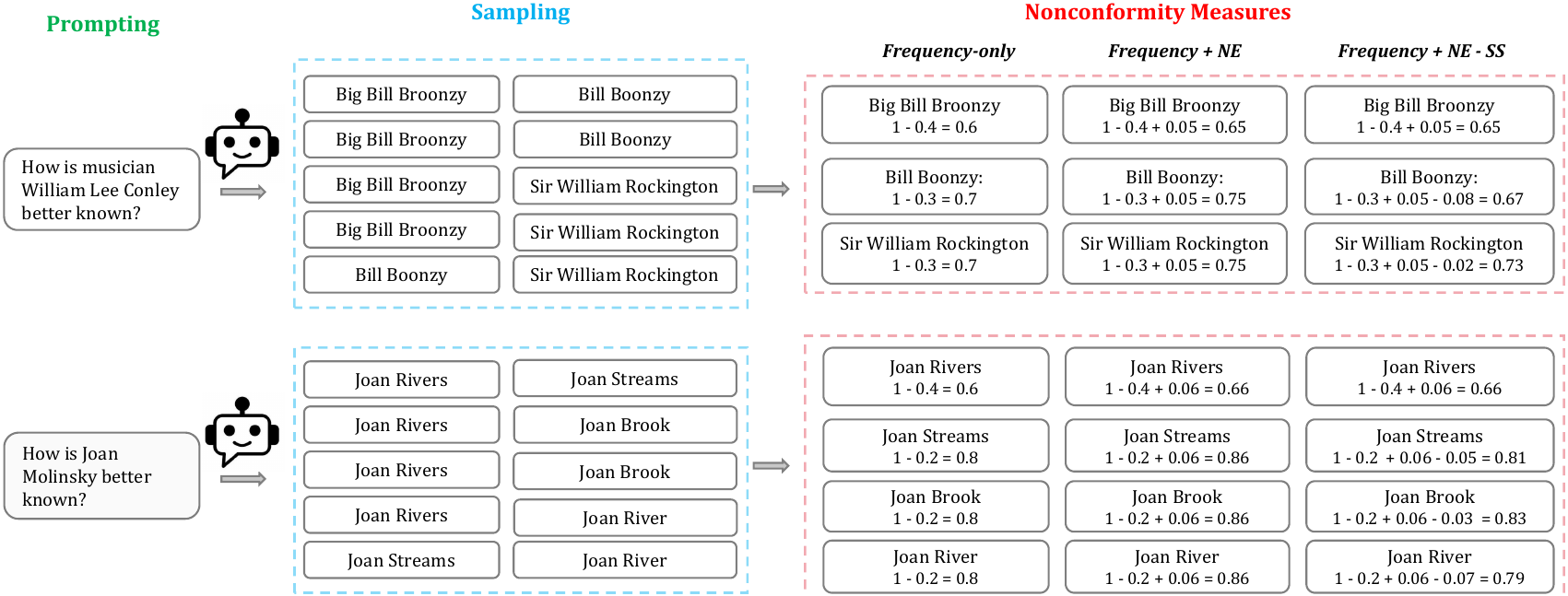}
    
\caption{\small Illustrations of the proposed problem and solution. Three uncertainty notions for measuring nonconformity: (1) Frequency-only, where the nonconformity score is calculated as $1-$ \textit{the frequency of a response out of 10 samplings}. Concentration issues arise at scores of \textbf{0.6, 0.7, and 0.8}. For instance, responses from different prompts (e.g., "Big Bill Broonzy" and "Joan Rivers") have the same score of 0.6, as well as responses within the same prompt (e.g., "Bill Boonzy" and "Sir William Rockington") which both have a score of 0.7, and so forth. (2) Frequency combined with NE, where the nonconformity score is calculated as $1-$ \textit{frequency} $+$ \textit{NE}, revealing concentration issues at scores of \textbf{0.75 and 0.86}. (3) Frequency, NE, and SS combined, where the nonconformity score is calculated as $1-$ \textit{frequency} $+$ \textit{NE} $-$ \textit{SS}, with \textbf{no observed concentration issues}.}
    \label{FIG:overview}
    \vspace{-3mm}
\end{figure*}


To enable CP without logit-access, a straightforward way is to calculate the frequency of each response via sampling and approximate model-based probabilities. However, we theoretically prove that this approach is extremely computationally expensive (Lemma \ref{lemma31}). As nonconformity scores typically measure the level of uncertainty, CP depends on the \textit{ranking} of the nonconformity measures rather than their actual \textit{values} \cite{shafer2008tutorial}. Therefore, we propose to sample responses for a certain number of times (e.g., 30) for each input and then utilize the frequency of each response as a \textit{coarse-grained} uncertainty notion. This approach reduces the overall sampling costs and eliminates the dependence on the logits. However, when using frequency as the only nonconformity measure, we observe that nonconformity scores concentrate on certain values as some responses may share the same frequency even if they have varied levels of uncertainty (see Figure \ref{FIG:overview}), consequently diminishing the efficiency of prediction sets. To distinguish between responses that share the same frequency, we first identify two potential causes: the respective concentration issues across different prompts and within the same prompt. We then propose two additional \textit{fine-grained} uncertainty notions: normalized entropy (NE), measuring prompt-wise self-consistency to alleviate concentration issues across different prompts; and semantic similarity (SS), measuring response-wise similarity to the most frequent response within the same prompt, to mitigate internal concentration issues specific to the prompt. Figure \ref{FIG:overview} illustrates the different nonconformity scores defined using \textit{frequency-only}, \textit{frequency combined with NE}, and \textit{frequency combined with NE and SS} as nonconformity measures, respectively. By considering various uncertainty information, the proposed non-conformity score function can better distinguish the uncertainty of different responses.

Our contributions are summarized as follows:
\begin{itemize}[topsep=0pt, before=\setlength{\itemsep}{0pt}, parsep=0pt, leftmargin=*]
\item To our knowledge, this is the first CP work dedicated to LLMs without logit-access that provides a coverage guarantee for the prediction set with a small size. 

\item We propose a novel CP approach that uses both course-grained and fine-grained uncertainty notions as the non-conformity measures. 
We also theoretically prove (1) it is computationally infeasible to use response frequency to approximate model output probability, and (2) our approach ensures a rigorous statistical coverage guarantee.

\item We conduct experiments on both close- and open-ended QA tasks and demonstrate the effectiveness of our method. Notably, we mostly surpass all baselines, including four logit-access methods and one method without logit-access.

\end{itemize}

\section{Preliminaries of Conformal Prediction}
Conformal prediction (CP) \cite{vovk2005algorithmic} is a model-agnostic method offering distribution-free uncertainty quantification, which produces prediction sets/intervals containing ground-truth labels with a desired error rate $\alpha$. One of the widely used CP methods is split CP. 
Formally, let $(X, Y)$ be a sample, where $X$ represents features and $Y$ represents the outcome. We denote the calibration set as $(X_i, Y_i)_{i=1,\ldots,n}$ and the test set as $(X_{\text{test}}, Y_{\text{test}})$. CP presents the following nesting property:
\begin{equation}
\small
\label{nesting-pro}
\alpha_1 > \alpha_2 \Rightarrow C_{1-\alpha_1}(X) \subseteq C_{1-\alpha_2}(X).
\end{equation}

\begin{theorem} [Conformal coverage guarantee]
\label{theorem21}
Suppose $(X_i, Y_i)_{i=1,...,n}$ and $(X_{\text{test}}, Y_{\text{test}})$ are independent and identically distributed (i.i.d.). $C_{1-\alpha}(X_{test})$ is a set-valued mapping satisfying the nesting property in Eq. \ref{nesting-pro}. Then the following holds: 
\begin{equation}
\small
\label{marginal}
    P(Y_{\text{test}} \in C_{1-\alpha}(X_{\text{test}})) \geq 1 - \alpha,
\end{equation}
where \(\alpha \in (0, 1)\) is the user-defined error rate.
\end{theorem}

\paragraph*{\textbf{Nonconformity Measures.}}
The nonconformity measure  \(N\) is a core element in CP. It measures uncertainty in the model's output by assessing the deviation of a specific instance or output from patterns observed in the training data. Typically, we have logit access to models to measure nonconformity, e.g., $1 - f(X)_Y$. For LLMs, \(N\) is typically derived from the post-hoc logits.



\paragraph*{\textbf{Split CP Steps.}}
\label{cp steps}
Split CP typically involves four steps \cite{angelopoulos2021gentle}:
\begin{enumerate}[before=\setlength{\itemsep}{0.1pt}, leftmargin=*]
\item Establish heuristic uncertainty notions. 
\item Define the nonconformity measures/score function $ N(x, y) \in \mathbb{R} $.
\label{step3}
\item Compute $ \hat{q} $ as the $  \frac{\lceil(n+1)(1-\alpha) \rceil}{n} $ quantile of the nonconformity scores.
\label{step4}
\item Use $ \hat{q} $ to generate prediction sets for new examples: $ C(X_{\text{test}}) = \{Y : N(X_{\text{test}}, Y) \leq \hat{q}\} $.
\end{enumerate}

\section{Methodology}
\label{motivation}
Our method considers two pivotal challenges arising from the LLMs without logit-access: how to approximate the logit information of LLMs; and how to further improve CP efficiency, i.e., small prediction sets. We propose the \textbf{Lo}git-\textbf{free} \textbf{C}onformal \textbf{P}rediction for LLMs (\textbf{LofreeCP}), where its nonconformity measures consist of three notions: \textit{frequency}, representing coarse-grained uncertainty; \textit{NE}, representing prompt-wise fine-grained uncertainty; and \textit{SS}, representing response-wise fine-grained uncertainty.
\subsection{Frequency As the Rankings Proxy}
\label{sec3.1}
A straightforward way is to approximate real predictive probabilities through a sufficiently large number of samplings. However, as we show in Lemma \ref{lemma31}, a minimum of 9,604 samples is required to achieve a 95\% confidence level with a 1\% margin of error. Therefore, the implementation is impractical due to computational constraints.
\begin{lemma}[Minimum Sample Size for Confident Probability Estimation]
\label{lemma31}
Let $freq(Y_i)$ be the frequency of outcome $Y_i$ in the sampling, $N_{\text{total}}$ be the total number of samplings, $p_i$ be the desired estimated probability, $\epsilon$ be the estimation error, and $\delta$ be the target confidence level. To determine the minimum sample size for confident probability estimation, for any given $\epsilon > 0$ and $0 < \delta < 1$, the following inequality must hold:
\begin{equation}
\small
\label{eq(3)}
   P\left\{\left|\frac{freq(Y_i)}{N_{\text{total}}} - p_i\right| \leq \epsilon\right\} \geq \delta.
\end{equation}
Then, the minimum sample size $N_{total}$ satisfying Inequality \ref{eq(3)} is given by:
\begin{equation}
\small
    N_{total} \geq \left(\frac{u_{1-(1-\delta)/2}}{2\epsilon}\right)^2,
\end{equation}
where $u_{1-(1-\delta)/2}$ is the quantile of the standard normal distribution corresponding to the confidence level $1-(1-\delta)/2$. The proof of Lemma \ref{lemma31} is given in Appendix \ref{proof-lemma31}.
\end{lemma}

\begin{figure}[t]
    \centering
    \begin{subfigure}[b]{1\linewidth}
        \centering
        \includegraphics[width=1\linewidth]{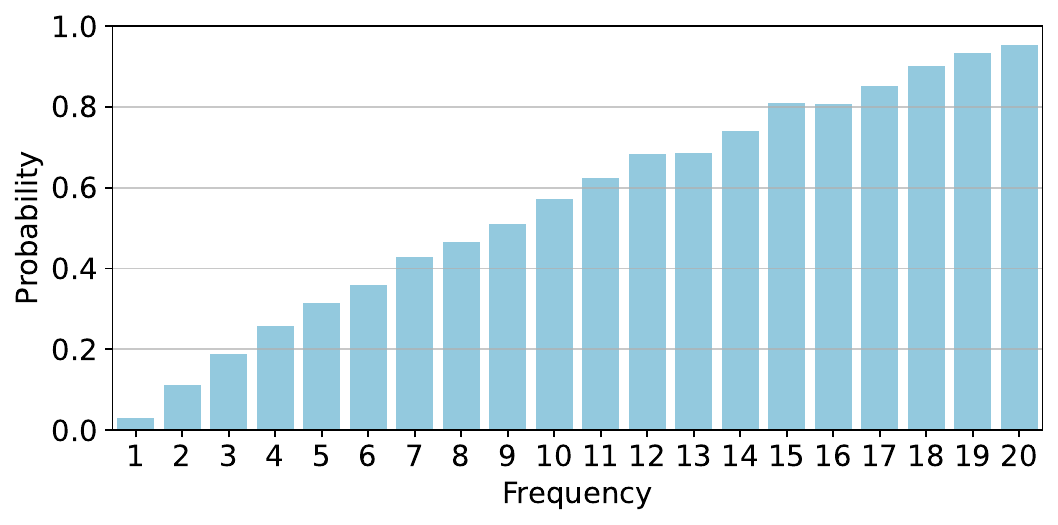}
        \caption{LLM response frequency vs. LLM output probability.}
        \label{FIG:1a}
    \end{subfigure}
    \begin{subfigure}[b]{1\linewidth}
        \centering
        \includegraphics[width=1\linewidth]{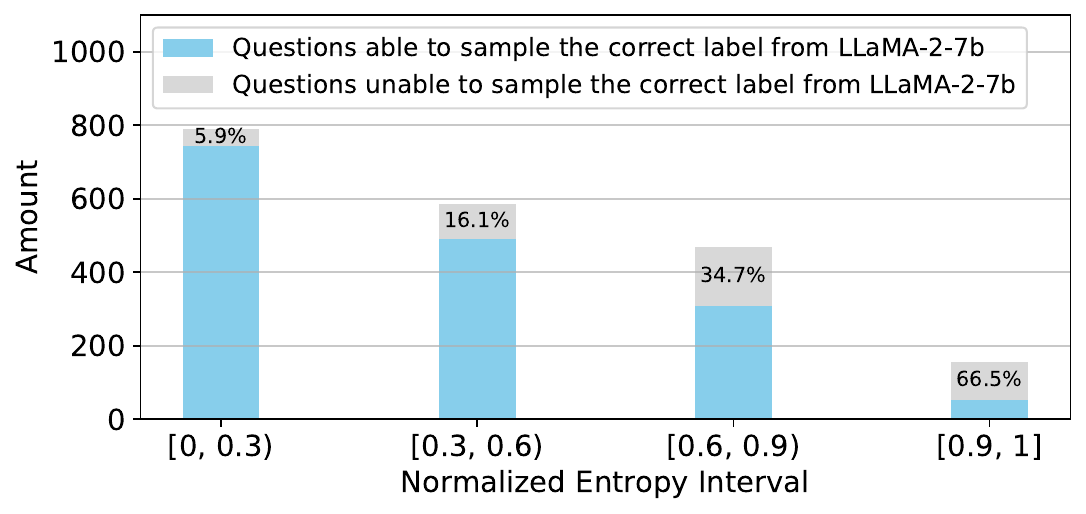}
        \caption{NE vs. the proportion of prompts unable to sample correct labels from Llama-2-7b.}
        \label{FIG:1b}
    \end{subfigure}
    \caption{\small Empirical findings with TriviaQA dataset. }
    \label{FIG:combined}
    \vspace{-2mm}
\end{figure}

Since nonconformity measures are grounded in assessing the model’s predictive uncertainty \cite{shafer2008tutorial}, the primary focus lies in the \textit{rankings} of uncertainty inherent in nonconformity measures rather than the absolute values themselves. Further, self-consistency theory \cite{wang2022self, li2022advance} states that a repetitively sampled response is viewed as a form of consistency linked to higher confidence in the response. To empirically validate this intuition, we randomly select 2000 questions from the TriviaQA dataset \cite{joshi2017triviaqa}. 
We conducted 20 samplings from the Llama-2-7b model \cite{touvron2023llama}, extracted logits, and subsequently computed model output probabilities. The observed results depicted in Figure \ref{FIG:1a} indicate a direct positive correlation between response frequency and average real probability. As the response frequency climbs, there is a corresponding increase in the average real probability, suggesting a growing level of confidence and certainty in the model's responses. Therefore, we propose to use frequency as the proxy of probability ranking. It is defined as 
\begin{equation}
\small
\label{freq_Eq}
    F(\hat{y}^{(i)}_{a}, m) = \frac{\tilde{p}[\hat{y}^{(i)}_{a}]}{m},
\end{equation}
where $\hat{y}^{(i)}_{a}$ is the $a$-th non-repeated sampled response for $i$-th prompt, $m$ is the sampling quantity from LLMs for each prompt.
However, only using response frequency as nonconformity measures results in the concentration of nonconformity scores on certain values. This issue makes it challenging to discern nonconformity differences among responses with the same scores, rendering ineffective calibration in CP. 

\subsection{Fine-grained Uncertainty Notions}
To resolve the concentration issue, we propose two fine-grained uncertainty measures. Firstly, inspired by self-consistency theory \cite{wang2022self, li2022advance}, we incorporate NE, a prompt-wise fine-grained uncertainty notion, to mitigate the concentration issue across different prompts. NE is a measure of the uncertainty or diversity in the model's predictions when generating responses to a given prompt. It is defined as 
\begin{equation}
\small
\label{ne_Eq}
    H(x^{(i)} | \{\hat{y}^{(i)}_j\}_{j=1}^m) = |\frac{\sum_{a=1}^{n} \tilde{F}(\hat{y}^{(i)}_a) \log (\tilde{F}(\hat{y}^{(i)}_a))}{\log m}|,
\end{equation}
where $x^{(i)}$ is the $i$-th instance of the prompt dataset, $m$ is the number of sampled responses, $n$ is the number of non-repeated responses, $\hat{y}^{(i)}_j$ is the $j$-th sampled response. Following experiments in Section \ref{sec3.1}, we show that as NE increases, the number of unanswered questions also increases (Figure \ref{FIG:1b}), indicating a rise in uncertainty. 

Secondly, to address concentration issues within a prompt, we introduce SS as a response-wise fine-grained uncertainty measure. This metric semantically assesses the similarity between each non-top-1 response and the top-1 response within a prompt. Intuitively, when two non-top-1 responses share the same frequency, the one more semantically similar to the top-1 response is more likely to express high confidence and low uncertainty. We use the cosine similarity to express SS. It is defined as
\begin{equation}
\small
\label{ss_Eq}
    SS(\hat{y}^{(i)}_{a}, P^{(i)}_{\text{highest}}) = \frac{\mathbf{v}(\hat{y}^{(i)}_{a}) \cdot \mathbf{v}(P^{(i)}_{\text{highest}})}{\|\mathbf{v}(\hat{y}^{(i)}_{a})\| \cdot \|\mathbf{v}(P^{(i)}_{\text{highest}})\|},
\end{equation}
where $\mathbf{v}(x)$ is the vector representation of $x$, $P^{(i)}_{\text{highest}}$ is the response having the highest frequency for $i$-th prompt. However, if the response to be measured is the one with the highest frequency, we do not consider SS with itself.

\subsection{CP for LLMs Without Logit-Access}
Considering both the coarse-grained and fine-grained uncertainty notions, the final nonconformity score function of LofreeCP is defined as
\begin{equation}
\small
\label{nonconformity-measures}
\begin{aligned}
    N^{(i)} &= -F(\hat{y}^{(i)}_{a}, m) + \lambda_1 \cdot H(x^{(i)} | \{\hat{y}^{(i)}_j\}_{j=1}^m) \\
    &\quad - \lambda_2 \cdot SS(\hat{y}^{(i)}_{a}, P^{(i)}_{\text{highest}}),
\end{aligned}
\end{equation}
where $\lambda = (\lambda_1, \lambda_2)$ representing a hyperparameter configuration controls the balance between the coarse-grained and fine-grained uncertainty notions. LofreeCP has the coverage guarantee:

\begin{proposition}[Coverage guarantee of LofreeCP]
\label{LofreeCP-guarantee}
    Suppose $(X_i, Y_i)_{i=1,\ldots,n}$ and $(X_{\text{test}}, Y_{\text{test}})$ are i.i.d. Let $C_{1-\alpha}(X_{\text{test}})$ be defined as in Step \ref{step4}. Then we have the coverage guarantee: 
\begin{equation}
\small
\label{LofreeCP-guarantee-equation}
    P\left\{Y_{\text{test}} \in C_{1-\alpha}\left(X_{\text{test}}\right)\right\} \geq 1 - \alpha,
    \nonumber
\end{equation}
    where $\alpha \in (0, 1)$ denotes the desired error rate. The proof of the coverage guarantee of LofreeCP is provided in Appendix \ref{proof-LofreeCP}.
\end{proposition}

LofreeCP consists of three stages: calibration, validation, and testing. The calibration stage aims to find the quantile based on the desired error rate. We sample $m$ responses from the LLM for each prompt and store them in a response pool. Then, we obtain the nonconformity scores of the true labels with the following rules: if the true label exists in the pool, we use the nonconformity measures from Equation \ref{nonconformity-measures} to calculate its nonconformity score; otherwise, we set the nonconformity score as $\infty$ to signify that it is nearly impossible to for the LLM to generate the true response. After obtaining all nonconformity scores of the calibration set, we find the quantile based on the desired error rate. We use this quantile as a threshold value for both the validation and test stages.

We then use the validation set to choose the optima hyperparameter configuration $\lambda = (\lambda_1, \lambda_2)$. Subsequently, we conduct evaluations on the test set using the chosen configuration. Both stages follow identical sampling steps to the calibration, traversing all responses and calculating the nonconformity scores. We preserve the responses whose nonconformity score is less than the threshold in our final prediction set. The pseudocode of the LofreeCP method is provided in Appendix \ref{pseudocode}.

\begin{table*}[t]
\caption{\small Results for TriviaQA using Llama-2-13b: Among all baselines, only \textit{First-K$_{\text{white}}$} and \textit{First-K$_{\text{black}}$} are non-CP-based, while the rest are CP-based methods. In the results, \textbf{bold} indicates that the method produces the best performance among all methods; \ding{55} denotes that the method fails to produce the set with the desired error rate.} 
\label{main-results-trivia}
\small
\centering

\begin{subtable}{\textwidth}
\centering
\scalebox{0.85}{%
\begin{tabular}{ccccc|ccc|cccc} 
\toprule[1pt] 
\multirow{3}{*}{Methods} & \multirow{3}{*}{\shortstack{Logit-Access}} & \multicolumn{9}{c}{Error Rate} \\
& &\multicolumn{3}{c}{0.2} & \multicolumn{3}{c}{0.25} & \multicolumn{3}{c}{0.3} \\
& & ECR & SSC$\uparrow$ & APSS$\downarrow$ & ECR & SSC$\uparrow$ & APSS$\downarrow$ & ECR & SSC$\uparrow$ & APSS$\downarrow$ \\
\midrule 
First-K$_{\text{white}}$ & \ding{51}& 82.1 & 76.6 & 3.39 & 76.1 & 72.9 & 1.90 & \ding{55} & \ding{55} & \ding{55} \\
CLM & \ding{51}  & 80.2 & 73.4 & 2.29 & 75.2 & 69.1 & 1.55 & 70.1 & 68.3 & 1.28\\
SCP& \ding{51}  & 80.3 & 75.7 & 2.25 & 75.1 & 70.0 & 1.59 & 70.3 & 74.5 & 1.21\\
SAPS & \ding{51}  & 80.0 & 77.9 & 2.74 & 75.1 & 64.2 & 1.80 & 70.0 & 49.4 & 1.55\\
First-K$_{\text{black}}$& {\ding{55}}  & 80.1 & 76.8 & 2.70 & 76.4 & 72.2 & 1.90 & \ding{55} & \ding{55} & \ding{55}\\
\textbf{LofreeCP (Ours)} &{\ding{55}}& 80.1 & \textbf{79.0} & \textbf{2.19} & 75.3 & \textbf{74.5} & \textbf{1.43} & 70.3 & \textbf{76.7} & \textbf{1.08}\\
\end{tabular}
}

\scalebox{0.85}{%
\begin{tabular}{ccccc|ccc|cccc} 
\toprule[0.65pt] 
\multirow{3}{*}{Methods} & \multirow{3}{*}{\shortstack{Logit-Access}} & \multicolumn{9}{c}{Error Rate} \\
& &\multicolumn{3}{c}{0.35} & \multicolumn{3}{c}{0.4} & \multicolumn{3}{c}{0.45} \\
& & ECR & SSC$\uparrow$ & APSS$\downarrow$ & ECR & SSC$\uparrow$ & APSS$\downarrow$ & ECR & SSC$\uparrow$ & APSS$\downarrow$ \\
\midrule 
First-K$_{\text{white}}$ & \ding{51}& \ding{55} & \ding{55} & \ding{55} & 62.4 & 62.5 & 1.00 & \ding{55} & \ding{55} & \ding{55} \\
CLM & \ding{51}  & 65.0 & 69.3 & 0.96 & 60.1 & 72.7 & 0.81 & 55.2 & 83.3 & 0.70\\
SCP& \ding{51}  & 65.1 & 76.4 & 1.02 & 60.3 & 75.7 & 0.85 & 55.3 & 82.5 & 0.74\\
SAPS& \ding{51}  & 65.1 & 57.4 & 1.28 & 60.1 & 70.7 & 0.85 & 55.1 & 76.5 & 0.72\\
First-K$_{\text{black}}$& {\ding{55}}  & 66.5 & 66.5 & 1.00 & \ding{55} & \ding{55} & \ding{55} & \ding{55} & \ding{55} & \ding{55}\\
\textbf{LofreeCP (Ours)} &{\ding{55}}& 65.1 & \textbf{78.5} & \textbf{0.90} & 60.0 & \textbf{81.0} & \textbf{0.75} & 55.2 & \textbf{84.1} & \textbf{0.66}\\
\bottomrule[1pt] 
\end{tabular}
}
\vspace{-0.3em} 
\end{subtable}
\end{table*}

\section{Experiments}

\subsection{Experimental Setup}
\paragraph*{\textbf{Backbone LLMs and Evaluation Tasks.}}
Since we need to compare LofreeCP with logit-based methods, from where logits can be retrieved directly, we consider different open-source LLMs, including Llama-2-7B, Llama-2-13B, WizardLM-v1.2(13b) \cite{xu2023wizardlm} and Vicuna-v1.5(7b) \cite{chiang2023vicuna} models as our backbone models. Note that our method uses these LLMs as if they were API-only LLMs, i.e., it assumes no access to any internal information of LLMs. We use both open-ended Question-Answering (QA) and close-ended Multi-Choice Question-Answering (MCQ) tasks for evaluation.


\paragraph*{\textbf{Datasets.}}
We use standard benchmarking datasets TriviaQA and MMLU \cite{hendrycks2020measuring}, following \cite{kumar2023conformal} and \cite{quach2023conformal}. We also include the WebQuestions benchmark \cite{berant2013semantic}. For QA, we use the TriviaQA dataset, which consists of trivia questions spanning a wide range of topics such as history and science, and the WebQuestions dataset, which is focused on questions asked by users on a search engine. MMLU dataset, covering 57 subjects (e.g., mathematics, history), is used for MCQ. We focus on a subset of 16 subjects out of the total 57, as in \citet{kumar2023conformal}.


\paragraph*{\textbf{Baselines.}} Baselines include methods without logit-access and those based on logit:
\begin{itemize}[topsep=0pt, before=\setlength{\itemsep}{0pt}, parsep=0pt, leftmargin=*]
\item \textbf{Top-K$_{\text{white}}$}. A logit-based non-CP method without coverage guarantee, which includes responses with the first $K$ highest probabilities for each prompt in the prediction set.
\item \textbf{Standard Split Conformal Prediction (SCP)} \cite{vovk2005algorithmic}. A logit-based CP method, which follows the steps shown in Section \ref{cp steps}.
\item \textbf{Sorted Adaptive Prediction Sets (SAPS)} \cite{huang2023conformal}. A logit-based CP method, which uses the highest probability and replaces other probabilities with some weighted values to mitigate the miscalibration issue.
\item \textbf{Top-K$_{\text{black}}$.} A non-CP method without logit-access and coverage guarantee, which includes responses with the first K highest frequency for each prompt in the prediction set.
\item \textbf{Conformal Language Modeling (CLM)} \cite{quach2023conformal}. The state-of-the-art logit-based CP method, which uses the general risk control framework. This baseline is only used in QA as it is not applied to MCQ.
\end{itemize}
\paragraph*{\textbf{Metrics.}} We use following metrics for evaluation \cite{angelopoulos2021gentle}:
\begin{itemize}[topsep=0pt, before=\setlength{\itemsep}{0pt}, parsep=0pt, leftmargin=*]
\item {Empirical Coverage Rate (ECR)} assesses whether the conformal procedure has the correct coverage with the theoretical guarantee.
\item {Size-Stratified Coverage (SSC) \cite{angelopoulos2020uncertainty}} assesses the worst coverage rate of each bin among different set sizes.


\item {Average Prediction Set Size (APSS)} assesses the efficiency of CP. We expect the APSS of an efficient CP method to be small.
\end{itemize}

\begin{table*}[h]
\caption{\small Results for WebQuestions using Llama-2-13b.}
\label{main-results-web}
\small
\centering
\begin{subtable}{\textwidth}
\centering
\scalebox{0.85}{%
\begin{tabular}{ccccc|ccc|ccc|cccc} 
\toprule[1pt]
\multirow{3}{*}{Methods} & \multirow{3}{*}{\shortstack{Logit-Access}} & \multicolumn{12}{c}{Error rate} \\
& &\multicolumn{3}{c}{0.35} & \multicolumn{3}{c}{0.4} & \multicolumn{3}{c}{0.45} & \multicolumn{3}{c}{0.5}\\
& & ECR & SSC$\uparrow$ & APSS$\downarrow$ & ECR & SSC$\uparrow$ & APSS$\downarrow$ & ECR & SSC$\uparrow$ & APSS$\downarrow$ & ECR & SSC$\uparrow$ & APSS$\downarrow$\\
\midrule 
{First-K$_{\text{white}}$} & \ding{51}& 66.4 & 57.5 & 6.18 & 61.6 & 58.1 & 3.81 & 57.5 & 55.0 & 2.91 & 50.6 & 49.0 & 1.97\\
CLM & \ding{51}  & 65.3 & 50.5 & \textbf{4.54} & 60.5 & 52.9 & 2.86 & 55.0 & 51.6 & 1.81 & 50.1 & 56.8 & 1.27\\
SCP& \ding{51}  & 65.1 & 46.7 & 4.61 & 61.6 & 49.3 & 3.01 & 55.2 & 55.8 & 2.02 & 50.2 & 57.8 & 1.39\\
SAPS& \ding{51}  & 65.2 & 46.2 & 5.19 & 60.6 & 56.2 & 3.39 & 55.5 & 37.7 & 2.40 & 50.8 & 21.7 & 1.86 \\
First-K$_{\text{black}}$& {\ding{55}}  & 65.1 & 54.9 & 6.20 & 60.0 & 55.3 & 3.78 & 56.9 & 54.4 & 2.91 & 53.7 & 52.4 & 1.97  \\
\textbf{LofreeCP (Ours)} &{\ding{55}}& 65.1 & \textbf{61.1} & 5.33 & 60.0 & \textbf{60.0} & \textbf{2.68} & 55.1 & \textbf{60.1} & \textbf{1.60} & 50.3 & \textbf{59.9} & \textbf{1.06}\\
\bottomrule[1pt] 
\end{tabular}
}
\end{subtable}
\vspace{-0.2em}
\end{table*}


\subsection{Results for QA} 
We perform QA using TriviaQA and WebQuestions datasets. The results for Llama-2-13b are reported in Tables \ref{main-results-trivia}-\ref{main-results-web}, those for Llama-2-7b are shown in the sensitivity analysis of Section \ref{model-size} and those for WizardLM-v1.2(13b) and Vicuna-7b-v1.5 can be found in Appendix \ref{add-results}. In Table \ref{main-results-trivia}, the LofreeCP method excels on TriviaQA across all error rate settings, outperforming the second-best method, CLM, by 7.7\% in terms of APSS at an error rate of 0.25. Regarding SSC, our LofreeCP method surpasses the second-best method, First-K$_{\text{white}}$, by 1.6\%. In Table \ref{main-results-web}, our method demonstrates superior performance on WebQuestions in most settings. For instance, at an error rate of 0.45, our LofreeCP method outperforms the second-best method, CLM, by 11.6\% in terms of APSS. Regarding SSC, we outperform the second-best method, SCP, by 4.3\%. WizardLM-v1.2(13b) and Vicuna-7b-v1.5 exhibit similar trends to Llama-2-13b.

The smallest APSS indicates that our method can produce the most efficient prediction sets. The highest SSC indicates that our method is attentive to the conditional coverage rate, achieving well-calibrated uncertainty estimates within diverse size categories. The rationale behind the observed superior performance is that our nonconformity measure can capture the coarse-grained uncertainty of responses and effectively optimize nonconformity through fine-grained considerations, thereby mitigating the inherent miscalibration issue in LLMs.



\subsection{Ablation Study}
\begin{figure}[t]
    \centering
    \includegraphics[width=0.47\textwidth]{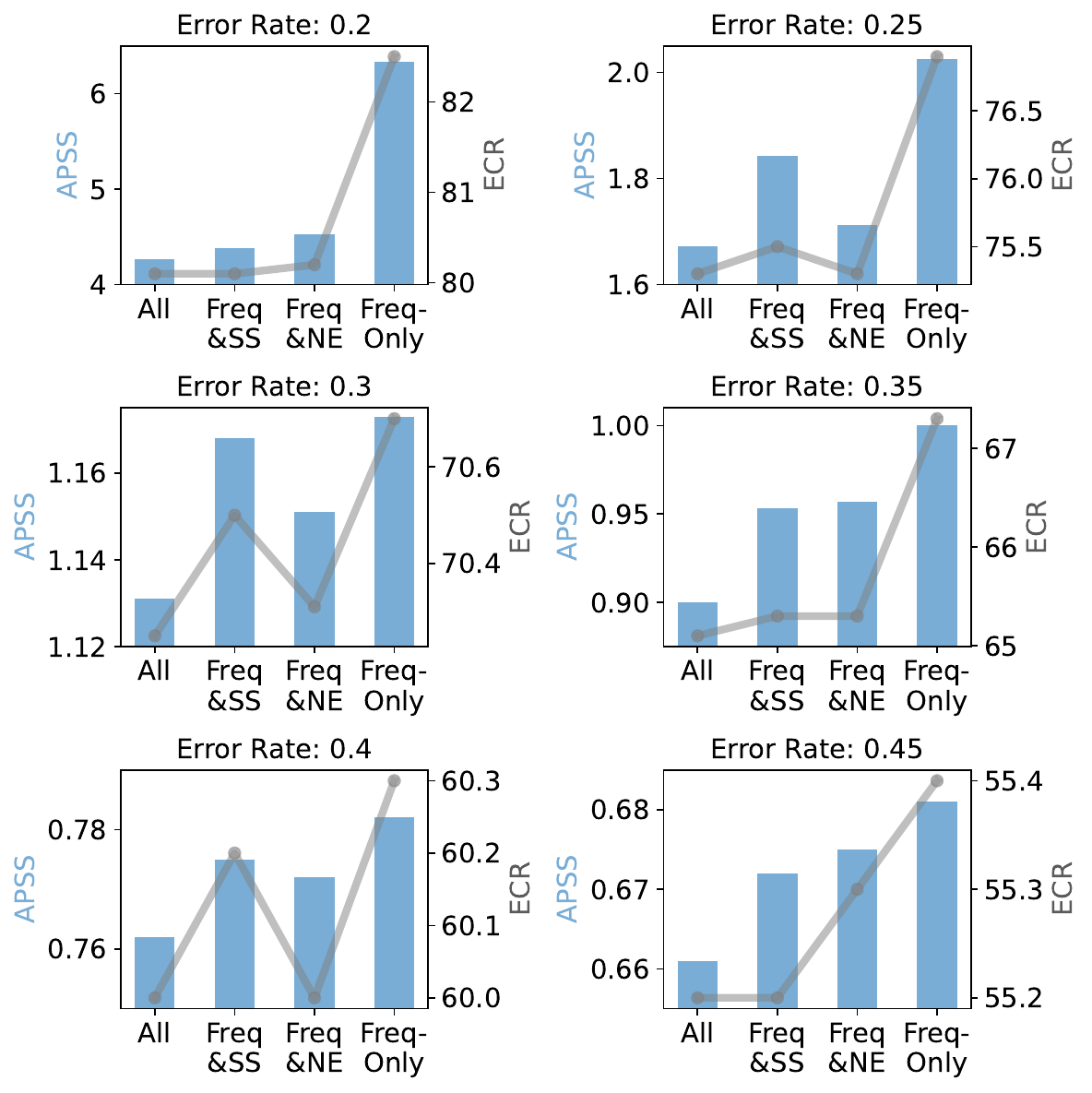}
    \caption{\small Ablation study. The \textcolor{blue}{blue} bar chart represents APSS, while the \textcolor{gray}{gray} line represents ECR.}
    \label{fig:ablation_study}
    \vspace{-2mm}
\end{figure}

To demonstrate the impact of our fine-grained uncertainty notions (NE and SS) on mitigating the concentration issues, we conduct a series of ablation studies using the TriviaQA dataset with a sampling quantity of 20. We compare LofreeCP with its different variants: we remove one fine-grained notion at a time (Freq\&SS, removing the NE notion; and Freq\&NE, removing the SS notion), and finally remove both fine-grained notions (Freq-Only). We report APSS and ECR, the direct indicators of the concentration issue, in Figure \ref{fig:ablation_study}.

\paragraph*{\textbf{Impact of Concentration Issue.}} As introduced in Section \ref{motivation}, the concentration issue occurs when the nonconformity score is concentrated on certain values. When we use the frequency-only variant (Freq-Only), this issue can be observed in all error rate settings, as shown in Figure \ref{fig:ablation_study}: Freq-Only has the largest APSS and the most conservative ECR. Due to its coarse-grained uncertainty notion, Freq-Only tends to generate similar non-conformity scores clustered into several groups, making it hard to differentiate granular uncertainties to produce efficient prediction sets. 

\paragraph*{\textbf{Full Method Mitigates Concentration Issue.}} We further observe that the concentration issue is mitigated in all error rate settings by incorporating fine-grained notions (NE \& SS). For example, at an error rate of 0.2, Freq-Only exhibits an APSS of nearly 6.5, while the full method LofreeCP has an APSS of 4.27, resulting in a drop of more than 23\%. The method including only SS or NE also mitigates the concentration issue to some extent, while the full method performs the best in terms of APSS and ECR. The results suggest that NE and SS both have a significant impact on improving the efficiency of prediction sets by mitigating concentration issues of nonconformity scores.

\begin{figure}[t]
    \centering
    \includegraphics[width=0.5\textwidth]{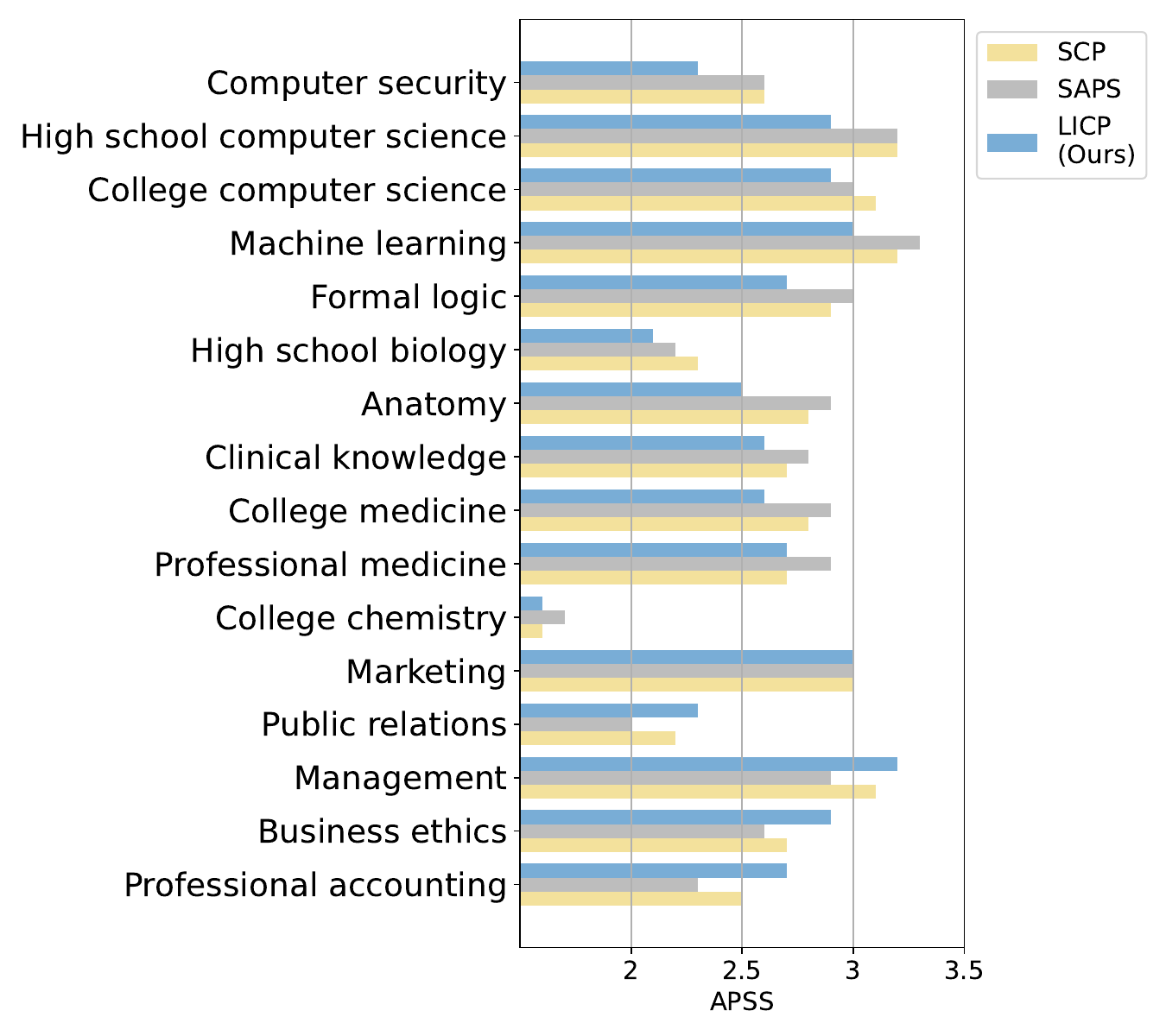}
    \caption{\small Results on MCQ task, with the error rate of 0.2. Our method and baselines are applied individually to each of the 16 subjects.} 
    \label{fig:5}
    \vspace{-2mm}
\end{figure}

\subsection{Results for MCQ}
In addition to open-ended tasks, e.g. QA, LofreeCP is also effective at close-ended tasks that can be converted into a generation pipeline, e.g. MCQ. We conduct MCQ experiments on the MMLU dataset using Llama-2-13b with a sampling quantity of 20. We present the results in Figure \ref{fig:5}.\footnote{We omit the results from top-K methods as they exhibit much larger APSS than other methods for MCQ.} LofreeCP exhibits superior performance. When compared with SCP and SAPS across all 16 subjects, LofreeCP achieves the best performance in 9 subjects and ties for the best in subjects of professional medicine, college chemistry, and marketing, resulting in the overall best performance in 12 out of 16 subjects. In contrast, SCP only ties for the best in 3 subjects. SAPS achieves the solo best performance in 3 subjects and ties for the best in 1 subject.


\begin{figure}[b]
    \centering
    \includegraphics[width=0.43\textwidth]{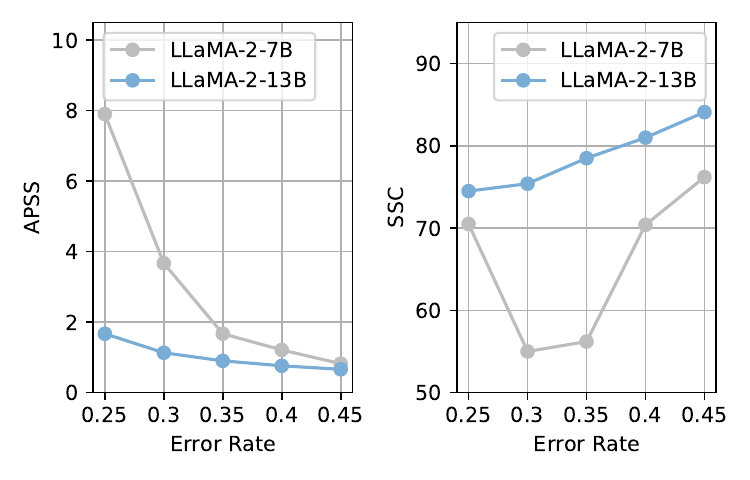}
    \caption{\small Results of the sensitivity analysis for different backbone models: Llama-2-7b and Llama-2-13b.}
    \label{fig:6}
    \vspace{-2mm}
\end{figure}

An intriguing observation is related to subjects in the business and management (B\&M) category (e.g., marketing and public relations). When using LofreeCP method, these subjects show slightly larger APSS than the two logit-based methods, SCP and SAPS. This suggests that the logits for responses to B\&M questions predicted by the Llama-2-13b model are better calibrated than the remaining subjects from the Science, Technology, Engineering, and Mathematics (STEM) category. Our LofreeCP method mitigates the model miscalibration issue by refraining from directly using logits. 

\subsection{Sensitivity Analyses}

\paragraph*{\textbf{BackBone Models.}}
\label{model-size}
To investigate the influence of different backbone models on the performance of LofreeCP, we conduct experiments using Llama-2-7b and Llama-2-13b with a sampling quantity of 20. Results of SSC and APSS are shown in Figure \ref{fig:6}. We observe that better performance of APSS and SSC in the 13b setting than in the 7b setting. We believe this is because Llama-2-13b is more powerful than Llama-2-7b, and produces more confident and calibrated responses, thereby providing more efficient prediction sets. Results for Vicuna-v1.5(7b) are provided in Appendix \ref{add-results}, indicating that Vicuna-v1.5(7b) can only produce prediction sets with higher error rates compared to Llama-2 backbones. This is because Vicuna-v1.5(7b) is less powerful for these two datasets. This demonstrates that CP performance for LLMs is largely dependent on the performance of the backbone models.

\begin{figure}[t]
    \centering
        \centering
        \includegraphics[width=\linewidth]{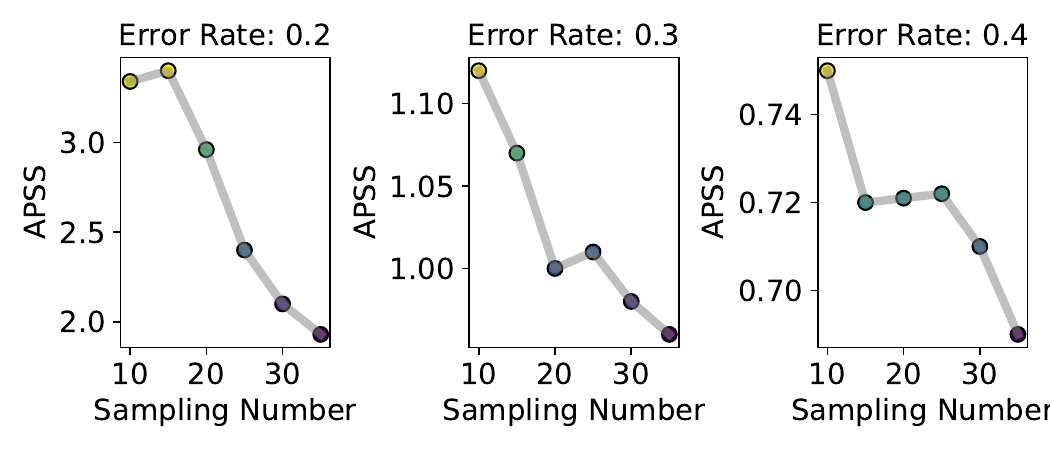}
        \caption{\small Sensitivity analysis of sampling quantity.}
        \label{FIG:sampling-sens}
    \vspace{-4mm}
\end{figure}

\paragraph*{\textbf{Sampling Quantity}}
The sampling quantity regulates the number and types of sampled responses acquired from LLMs, thereby influencing frequency, NE and SS. We vary the sampling quantity from 10 to 40 on the TriviaQA dataset using Llama-2-13b, incrementing by 5 each time. Results shown in Figure \ref{FIG:sampling-sens} suggest that a larger sampling quantity tends to present better performance w.r.t. efficiency. This is because, with a higher sampling quantity, the frequency notion more accurately represents response rankings. Of particular interest is that, at an error rate of 0.2, the sampling quantity of 15 exhibits inferior performance compared to the quantity of 10. We hypothesize it is because a sampling quantity of 15 remains insufficient to adequately represent rankings meanwhile introducing more non-robust randomness in responses. In addition, we observe a larger impact of the sampling quantity on APSS when a small error rate guarantee is required. 

\begin{figure}[t]
        \centering
        \includegraphics[width=\linewidth]{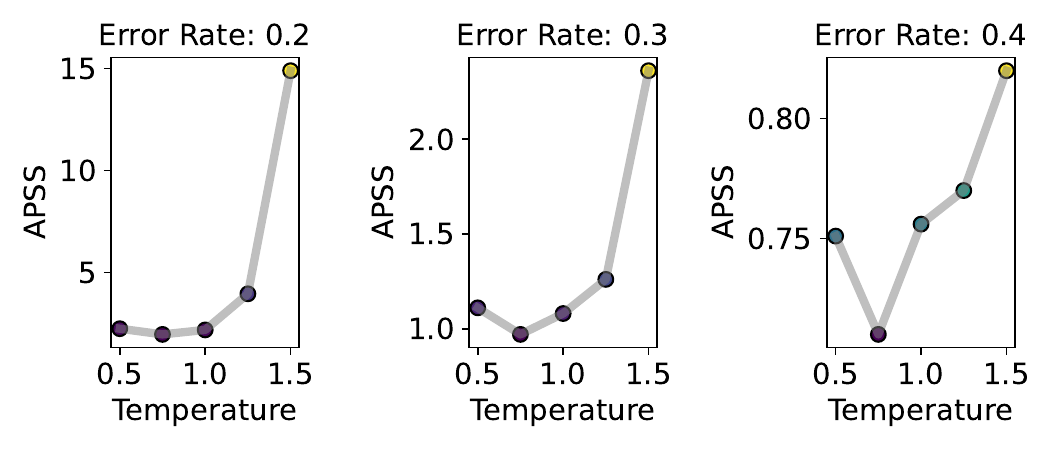}
        \caption{\small Sensitivity analysis of temperature.}
        \label{FIG:temp-ses}
    \vspace{-2mm}
\end{figure}

\paragraph*{\textbf{Temperature Scaling.}}
The temperature \cite{hinton2015distilling} in LLMs adjusts the randomness in generated outputs by scaling logits during the softmax operation. Higher temperatures boost the diversity of the output, which may further affect the performance of LofreeCP. In this experiment, we vary temperatures\footnote{ Temperature ranges between 0 and 2.} (0.5, 0.75, 1.0, 1.25, and 1.5) in the Llama-2-13b model. Results for the TriviaQA dataset are presented in Figure \ref{FIG:temp-ses}. The smallest (best) APSS is observed at a temperature of 0.75. We observe an overall growing trend as the temperature increases from 0.75 to 1.50. This indicates that excessive diversity can result in uncertain and suboptimal predictions. The decline from 0.50 to 0.75 implies that too much determinism may hurt CP efficiency due to a lack of randomness and diversity. We also note a significant temperature influence on APSS when aiming for low error rates. 

\section{Related Work}

\paragraph*{Conformal Prediction for NLP.} CP has already found diverse applications in NLP, e.g., text infilling and part-of-speech prediction \citet{dey2021conformal}, sentiment analysis \citet{pmlr-v128-maltoudoglou20a}, and Automatic Speech Recognition \citet{pmlr-v204-ernez23a}. In the application of CP to LLMs, existing methods are predominantly logit-based. For instance, \citet{kumar2023conformal} apply standard CP \cite{vovk2005algorithmic} to Llama-2-13b \cite{touvron2023llama} for the MCQ task by computing softmax scores of token logits for options to measure nonconformity. Similarly, \citet{quach2023conformal} extend CP to LLMs using the general risk control framework \cite{angelopoulos2021learn}. However, recent studies have pointed out that relying solely on logits may be flawed due to the potential issue of hallucinations in LLMs~\citep{lecun2023large}. Consequently, there is ongoing research aiming to reduce reliance on logits. \citet{huang2023conformal} propose to use the highest probability and replace other probabilities with weighted values. All these methods involve the utilization of logits.

\paragraph*{Uncertainty Estimation in LLMs.}Recent developments in LLMs have highlighted the importance of estimating their uncertainty. While there has been significant research on uncertainty in NLP~\citep{9761166, ulmer-etal-2022-exploring}, several methods exist to estimate the confidence of LLMs. These include Deep Ensemble methods~\citep{NIPS2017_9ef2ed4b}, Monte Carlo dropout~\citep{pmlr-v48-gal16}, Density-based estimation~\citep{yoo2022detection}, Confidence learning~\citep{devries2018learning}, as well as approaches based on logits. However, recent studies highlight concerns that LLMs may generate unfaithful and nonfactual content~\citep{maynez-etal-2020-faithfulness}. Additionally, the logits of LLMs' outputs often exhibit overconfidence when producing these incorrect answers, indicating that logits alone may not be entirely reliable for studying uncertainty \citep{desai2020calibration, miao2021prevent, vasconcelos2023generation}.
\section{Conclusion}

We study the critical problem of CP for API-only LLMs without logit-access. We propose a novel solution to define the nonconformity score function by leveraging uncertainty information from diverse sources. In particular, under a limited sampling budget, we first use the response frequency as the coarse-grained proxy of uncertainty levels. We then propose two fine-grained uncertainty notions (NE and SS) to further distinguish uncertainty at a nuanced level. Our proposed approach does not rely on model logits and can alleviate the known miscalibration issue when using logits. Experiments demonstrate the superior performance of our approach compared to logit-based and logit-free baselines. Our work opens up a new avenue to uncertainty estimation in LLMs without logit-access. 
\section*{Limitations}





Our approach encounters a common limitation of open-ended Natural Language Generation (NLG) tasks: the unbounded output space. In our work, we address this challenge by sampling a fixed number of times for every prompt from LLMs to achieve a comprehensive output space, but we recognize the potential for more effective and convincing approaches to handle this issue within the framework of CP. Secondly, another future direction is to expand our CP method to non-exchangeability scenarios, particularly in NLG domains, where calibration and test sets may not adhere strictly to the assumption of being independent and identically distributed (i.i.d.). Finally, due to financial constraints, we do not evaluate our approach on several proprietary LLMs (e.g., GPT 4) that allow users to obtain token log probabilities. Thus future work can validate our method on these models.

\bibliography{mypaper}
\bibliographystyle{acl_natbib}

\appendix
\section{Theoretical Proofs}
\subsection{Proof of Lemma \ref{lemma31}}
\label{proof-lemma31}
\begin{proof}[Proof]
When $N_{total}$ is sufficiently large, the Lindeberg–Lévy central limit theorem yields the following equation:
\begin{align*}
    \frac{\frac{freq(Y_i)}{N_{total}}-p_i}{\sqrt{p_i(1-p_i)/N_{total}}}\sim N(0,1).
\end{align*}
From this, we conclude that
\begin{align*}
    P\left\{\left|\frac{\frac{freq(Y_i)}{N_{total}}-p_i}{\sqrt{p_i(1-p_i)/N_{total}}}\right| \leq u_{1-(1-\delta)/2}\right\}=\delta.
\end{align*}
Approximately replacing $p_i$ in the denominator with $\frac{\text{freq}(Y_i)}{N_{total}}$, we obtain
\begin{align*}
    &P\{\frac{freq(Y_i)}{N_{total}}-u_{1-(1-\delta)/2}\cdot \\& \sqrt{\frac{\frac{freq(Y_i)}{N_{total}}(1-\frac{freq(Y_i)}{N_{total}})}{N_{total}}}\leq\\
    &p_i\leq \frac{freq(Y_i)}{N_{total}}+u_{1-(1-\delta)/2}\cdot \\& \sqrt{\frac{\frac{freq(Y_i)}{N_{total}}(1-\frac{freq(Y_i)}{N_{total}})}{N_{total}}}\}\\
    &=\delta.
\end{align*}
Therefore, to ensure
\begin{align*}
    u_{1-(1-\delta)/2}\cdot\sqrt{\frac{\frac{freq(Y_i)}{N_{total}}(1-\frac{freq(Y_i)}{N_{total}})}{N_{total}}}\cdot 2\leq 2\epsilon,
\end{align*}
we only need
\begin{align*}
    \sqrt{\frac{1/4}{N_{total}}}\cdot u_{1-(1-\delta)/2}\cdot2\leq 2\epsilon.
\end{align*}
This simplifies to
\begin{align*}
    N_{total} \geq \left(\frac{u_{1-(1-\delta)/2}}{2\epsilon}\right)^2 
\end{align*}
\end{proof}

\subsection{Proof of Proposition \ref{LofreeCP-guarantee}}
\label{proof-LofreeCP}
\begin{proof}[Proof]
Let $N$ denote the nonconformity measures of the calibration set $(X_i, Y_i)_{i=1,\ldots,n}$, and let ${\alpha}_1$ and ${\alpha}_2$ be the desired error rates, where ${\alpha}_1 > {\alpha}_2$. As indicated in Step \ref{step3}, we have $\hat{q_1} \leq \hat{q_2}$. Given $C(X_{\text{test}}) = \{Y : N(X_{\text{test}}, Y) \leq \hat{q}\}$, it follows that $C_{1-{\alpha}_1}(X) \subseteq C_{1-{\alpha}_2}(X)$. Consequently, the nesting property, as defined in Equation \ref{nesting-pro}, is satisfied. Therefore, Proposition \ref{LofreeCP-guarantee} holds.
\end{proof}

\section{Implementation Details}
\label{details}
\subsection{Dataset}
The TriviaQA benchmark (available at \url{https://nlp.cs.washington.edu/triviaqa/} or can be accessed from Hugging Face at \url{https://huggingface.co/datasets/trivia\_qa}) and the WebQuestions benchmark (available at \url{worksheets.codalab.org} or can be accessed from Hugging Face at \url{https://huggingface.co/datasets/web\_questions}) are employed for QA. Both datasets operate within a closed-book setting, where LLMs refrain from using supporting text when answering questions. 

The MMLU benchmark (can be accessed from Hugging Face at \url{https://huggingface.co/datasets/lukaemon/mmlu}) is designed for MCQ, which covers 57 subjects across STEM, the humanities, the social sciences, and more. For our MCQ experiments, we leverage the dataset containing 16 subjects from the MMLU benchmark: computer security, high school computer science, college computer science, machine learning, formal logic, high school biology, anatomy, clinical knowledge, college medicine, professional medicine, college chemistry, marketing, public relations, management, business ethics, professional accounting. 

For the TriviaQA dataset, we randomly select 10,000 question-answer pairs. Similarly, for the WebQuestions dataset, we randomly select 5,000 question-answer pairs. Regarding the MMLU dataset, we use all available data for each of the 16 subjects. Across all three datasets, we apply the same splitting strategy: 50\% of the data serves as the calibration set, 25\% as the validation set, and 25\% as the test set for each trial.

\subsection{Backbone LLMs}
We utilize the Hugging Face API to access open-source LLMs in our experiments, including Llama-2-7B (accessible at \url{huggingface.co/meta-llama/Llama-2-7b-hf}), Llama-2-13B (accessible at \url{huggingface.co/meta-llama/Llama-2-13b-hf}), WizardLM-v1.2(13b) (accessible at \url{huggingface.co/WizardLM/WizardLM-13B-V1.2}), and Vicuna-v1.5(7b) (accessible at \url{huggingface.co/lmsys/vicuna-7b-v1.5}). Access to Llama-2-7b and Llama-2-13b requires requesting approval via the Meta website (\url{https://llama.meta.com/}). Upon approval, access to these resources will be granted.

\subsection{Length-Normalization}
We use length normalization \cite{wu2016google} on logits to obtain response probability/likelihood:
\begin{align*}
   p(x, y_{k}) = \text{exp}(\frac{\text{log}p_{\theta}(y_k|x)}{lp(y_k)})
\end{align*}
where
\begin{align*}
   lp(y) = \frac{(5 + |y|)^{0.6}}{(5+1)^{0.6}}
\end{align*}

\subsection{Evaluation}
We extract an answer by analyzing the text until we encounter the first line break, comma, or period. This implies that in the dataset, we will disregard data whose answers contain line breaks, commas, or periods. Following this, we standardize the answers by converting them to lowercase, removing articles, punctuation, and duplicate whitespace. The generated answers are then evaluated using the exact match metric, where an answer is considered correct only if it exactly matches the provided answer. These guidelines align with those described in \citet{quach2023conformal}.

For SSC, We focus exclusively on bins with a set size greater than 0 and a sample number exceeding 10\% of the total test samples. This is because bins with a size of 0 and fewer samples lack reliability for coverage measurement.

\subsection{LLMs Parameters} We employ the default Transformer generative LMs parameters for our experiments, using default standard sampling with do\_sample set to True, top\_k set to 0, top\_p set to 1, and Temperature set to 1, except when conducting model hyperparameter-tuning experiments. In such hyperparameter-tuning cases, we explicitly mention the parameters in main body of the paper.

\subsection{Semantic Similarity}

The measure of semantic similarity was established leveraging the FastText model available within the gensim package. The configuration parameters were carefully selected, defining a vector size of 200 and imposing a minimum count threshold of 1 to ensure robustness and inclusivity in the model's representations.

\subsection{Experiment trails} We conduct 50 trials for all experiments, then average the results to eliminate randomness during the calibration.

\subsection{Error Rate Settings}
We do not apply the same error rate settings across different models or datasets. This is because each model varies in its coverage ability for the same dataset. Likewise, the same model doesn't possess identical coverage abilities for different datasets. Therefore, we adjust error rate settings for different combinations of model and dataset accordingly.

\subsection{GPUs}
We utilize six NVIDIA RTX 3090 graphics cards to support experiments.

\subsection{Pseudocode}
We show the pseudocode in Method \ref{pseudocode-details}, where we do not explicitly display the repetitive process of using various hyperparameter configurations to determine the best one. In our actual implementations,  we explore the range [0:0.05:2] for both $\lambda_1$ and $\lambda_2$. This range spans from 0 to 2, with each step incrementing by 0.05, thus covering values such as 0, 0.05, 0.1, 0.15, and so forth up to 2. Subsequently, we form different combinations to execute the calibration and validation stages. Ultimately, we utilize the best hyperparameter configurations for testing purposes.

\label{pseudocode}
\begin{method}[h]
\small
\caption{LofreeCP method}
\label{pseudocode-details}
\begin{algorithmic}[1]
    \Require Prompt $x^{(i)}$, LLM $f_{\theta}$, response $\hat{y}^{(i)}_j$, current sampling number $j$, required sampling number $m$, response pool $P^{(i)}$, response with the highest frequency $P^{(i)}_{highest}$, semantic similarity between response $a$ and $b$: $S($a$, $b$)$
    \For{$x^{(i)}$, $i = 1$ to $n$} 
    
    $P^{(i)} = \{\}$ 
    \Comment{\textcolor{blue}{\textbf{Calibration stage starts}}}
        \For{$j = 1$ to $m$} 
        
        $\hat{y}^{(i)}_j \xleftarrow{} f_{\theta}(x^{(i)})$ 
        \Comment{\textcolor{RoyalBlue}{Sample response from LLM given the prompt}}
            \If{$\hat{y}^{(i)}_j$ in $P^{(i)}$} 
            
            $\tilde{p}[\hat{y}^{(i)}_j]$ ++ 
                \Comment{\textcolor{RoyalBlue}{Increment frequency for existing response}}
            \Else 
            
            $\ \tilde{p}[\hat{y}^{(i)}_j] = 1$ 
            \Comment{\textcolor{RoyalBlue}{Initialize frequency for new response}}
            \EndIf
        \EndFor
        \State Sort($P^{(i)}$)
        \State Get $P^{(i)}_{highest}$
        \Comment{\textcolor{RoyalBlue}{Get the response with the highest frequency}}
        \label{step8}
        \If{$y^{(i)}$ in $P^{(i)}$} 
        
        $\ N^{(i)} = \frac{\tilde{p}[\hat{y}^{(i)}_{a}]}{m}$ + $\lambda_1$ $\cdot$ $H(x^{(i)} | \{\hat{y}^{(i)}_j\}_{j=1}^m)$ - $\lambda_2$ $\cdot$ $S(\hat{y}^{(i)}_{a}, P^{(i)}_{highest})$
        \Else 
        
        $\ N^{(i)} = \infty$ 
        \Comment{\textcolor{RoyalBlue}{Nonconformity measures}}
            
        \EndIf

    \EndFor
    \State $\hat{q}_{\alpha}$ = Quantile$(\{{N^{(1)}}, {N^{(2)}}, ..., {N^{(n)}}\}, \frac{\lceil(n+1)(1-\alpha)\rceil}{n})$
    \Comment{\textcolor{RoyalBlue}{Find quantile $\hat{q}_{\alpha}$}} 
    \Comment{\textcolor{blue}{\textbf{Calibration stage ends}}}
    \For{sampling same as 1 \texttt{\char`~} 7}
    \Comment{\textcolor{blue}{\textbf{Validation / Test stage starts}}}
    \For{each $\hat{y}^{(i)}_\alpha$ in $P^{(i)}$}
         
         $N^{(i)}_{\alpha}$ = $\frac{P^{(i)}[\hat{y}^{(i)}_{\alpha}]}{m}$ + $\lambda_1$ $\cdot$ $H(x^{(i)} | \{\hat{y}^{(i)}_j\}_{j=1}^m)$ - $\lambda_2$ $\cdot$ $S(\hat{y}^{(i)}_{\alpha}, P^{(i)}_{highest})$
    \EndFor
        \State $ C(x^{(i)}_{\text{test}}) = \{\hat{y}^{(i)}_{a} : N^{(i)}_{\alpha} \leq \hat{q}\} $ 
   \Comment{\textcolor{RoyalBlue}{Nonconformity measures}}
   \EndFor 
   \Comment{\textcolor{blue}{\textbf{Validation / Test stage ends}}
\end{algorithmic}
}
\end{method}

\section{Prompts}
\subsection{Few-shot Prompts of TriviaQA}
We use the 32-shot question-answer pair prompts from the TriviaQA dev set, the same as those in \citet{quach2023conformal}.

\begin{flushleft}
    {\ttfamily
    Answer these questions. \\
    Q: Which American-born Sinclair won the Nobel Prize for Literature in 1930? \\
    A: Sinclair Lewis \\
    
    Q: Where in England was Dame Judi Dench born? \\
    A: York \\
    
    Q: In which decade did Billboard magazine first publish an American hit chart? \\
    A: 30s \\
    
    Q: From which country did Angola achieve independence in 1975? \\
    A: Portugal \\
    
    Q: Which city does David Soul come from? \\
    A: Chicago \\
    
    Q: Who won Super Bowl XX? \\
    A: Chicago Bears \\
    
    Q: Which was the first European country to abolish capital punishment? \\
    A: Norway \\
    
    Q: In which country did the widespread use of ISDN begin in 1988? \\
    A: Japan \\
    
    Q: What is Bruce Willis’ real first name? \\
    A: Walter \\
    
    Q: Which William wrote the novel Lord Of The Flies? \\
    A: Golding \\
    
    Q: Which innovation for the car was developed by Prince Henry of Prussia in 1911? \\
    A: Windshield wipers \\
    
    Q: How is musician William Lee Conley better known? \\
    A: Big Bill Broonzy \\
    
    Q: How is Joan Molinsky better known? \\
    A: Joan Rivers \\
    
    Q: In which branch of the arts is Patricia Neary famous? \\
    A: Ballet \\
    
    Q: Which country is Europe’s largest silk producer? \\
    A: Italy \\
    
    Q: The VS-300 was a type of what? \\
    A: Helicopter \\
    
    Q: At which university did Joseph Goebbels become a doctor of philosophy? \\
    A: Heidelberg \\
    
    Q: Which prince is Queen Elizabeth II’s youngest son? \\
    A: Edward \\
    
    Q: When did the founder of Jehovah’s Witnesses say the world would end? \\
    A: 1914 \\
    
    Q: Who found the remains of the Titanic? \\
    A: Robert Ballard \\
    
    Q: Who was the only Spice Girl not to have a middle name? \\
    A: Posh Spice \\
    
    Q: What are the international registration letters of a vehicle from Algeria? \\
    A: DZ \\
    
    Q: How did Jock die in Dallas? \\
    A: Helicopter accident \\
    
    Q: What star sign is Michael Caine? \\
    A: Pisces \\
    
    Q: Who wrote the novel Evening Class? \\
    A: Maeve Binchy \\
    
    Q: Which country does the airline Air Pacific come from? \\
    A: Fiji \\
    
    Q: In which branch of the arts does Allegra Kent work? \\
    A: Ballet \\
    
    Q: Banting and Best pioneered the use of what? \\
    A: Insulin \\
    
    Q: Who directed the movie La Dolce Vita? \\
    A: Federico Fellini \\
    
    Q: Which country does the airline LACSA come from? \\
    A: Costa Rica \\
    
    Q: Who directed 2001: A Space Odyssey? \\
    A: Stanley Kubrick \\
    
    Q: Which is the largest of the Japanese Volcano Islands? \\
    A: Iwo Jima \\

    Q: (Question) \\
    A:
        }
\end{flushleft}

\subsection{Prompts of Webquestions}
We also use 32-shot question-answer pair prompts from the Webquestions train set.
\begin{flushleft}
    {\ttfamily
    Answer these questions. \\
    Q: What country is the Grand Bahama Island in? \\
    A: Bahamas \\
    
    Q: What two countries invaded Poland in the beginning of WW2? \\
    A: Germany \\
    
    Q: Which countries border the US? \\
    A: Canada \\
    
    Q: Where is Rome, Italy located on a map? \\
    A: Rome \\
    
    Q: What is Nina Dobrev's nationality? \\
    A: Bulgaria \\
    
    Q: What country does Iceland belong to? \\
    A: Iceland \\
    
    Q: What does Thai mean? \\
    A: Language \\
    
    Q: Who was Ishmael's mom? \\
    A: Hagar \\
    
    Q: What are the major cities in France? \\
    A: Paris \\
    
    Q: What city did Esther live in? \\
    A: Susa \\
    
    Q: What sport do the Toronto Maple Leafs play? \\
    A: Ice Hockey \\
    
    Q: What is Martin Cooper doing now? \\
    A: Inventor \\
    
    Q: What county is the city of Hampton, VA in? \\
    A: Hampton \\
    
    Q: What county is Heathrow Airport in? \\
    A: London \\
    
    Q: What type of car does Michael Weston drive? \\
    A: Wishcraft \\
    
    Q: What was Tupac's name in Juice? \\
    A: Bishop \\
    
    Q: Who does Maggie Grace play in Taken? \\
    A: Kim \\
    
    Q: What style of music did Louis Armstrong play? \\
    A: Jazz \\
    
    Q: Where does Jackie French live? \\
    A: Australia \\
    
    Q: Where is Jack Daniels factory? \\
    A: Tennessee \\
    
    Q: What is Charles Darwin famous for? \\
    A: Evolution \\
    
    Q: Where to visit in N. Ireland? \\
    A: Antrim \\
    
    Q: What are dollars called in Spain? \\
    A: Peseta \\
    
    Q: Who plays Meg in Family Guy? \\
    A: Mila Kunis \\
    
    Q: Where did Martin Luther King get shot? \\
    A: Memphis \\
    
    Q: What was Nelson Mandela's religion? \\
    A: Methodism \\
    
    Q: Who will win the 2011 NHL Stanley Cup? \\
    A: Canada \\
    
    Q: What is Henry Clay known for? \\
    A: Lawyer \\
    
    Q: What is the money of Spain called? \\
    A: Euro \\
    
    Q: Where are Sunbeam microwaves made? \\
    A: Florida \\
    
    Q: Where was Kennedy when he got shot? \\
    A: Dallas \\
    
    Q: Where did the Casey Anthony case take place? \\
    A: Orlando \\

    Q: (Question) \\
    A:
    }
\end{flushleft}

\subsection{Prompts of MMLU}
Each subject in MMLU uses similar prompts. We take the high school biology as examples.
\begin{flushleft}
    {\ttfamily
    Please engage in the multiple-choice question-answering task. You should generate the option (A, B, C, or D) you think is right. Examples are provided. \\
    (Select 8-shot randomly from other subjects)\\
    \vspace{0.3cm}
    This is a question from high school biology. \\
    A piece of potato is dropped into a beaker of pure water. Which of the following describes the activity after the potato is immersed into the water? \\
    (A) Water moves from the potato into the surrounding water. \\
    (B) Water moves from the surrounding water into the potato. \\
    (C) Potato cells plasmolyze. \\
    (D) Solutes in the water move into the potato.\\
    The correct answer is option: B.\\
    \vspace{0.3cm}
    You are the world’s best expert in high school biology. Reason step-by-step and answer the following question. \\ 
    From the solubility rules, which of the following is true?\\
    (A) All chlorides, bromides, and iodides are soluble \\
    (B) All sulfates are soluble \\
    (C) All hydroxides are soluble \\
    (D) All ammonium-containing compounds are soluble\\
    The correct answer is option:
    }
\end{flushleft}

\section{Additional Results}
\label{add-results}

\subsection{Sensitivity Experiments}
\begin{figure}[h]
    \centering
        \centering
        \includegraphics[width=\linewidth]{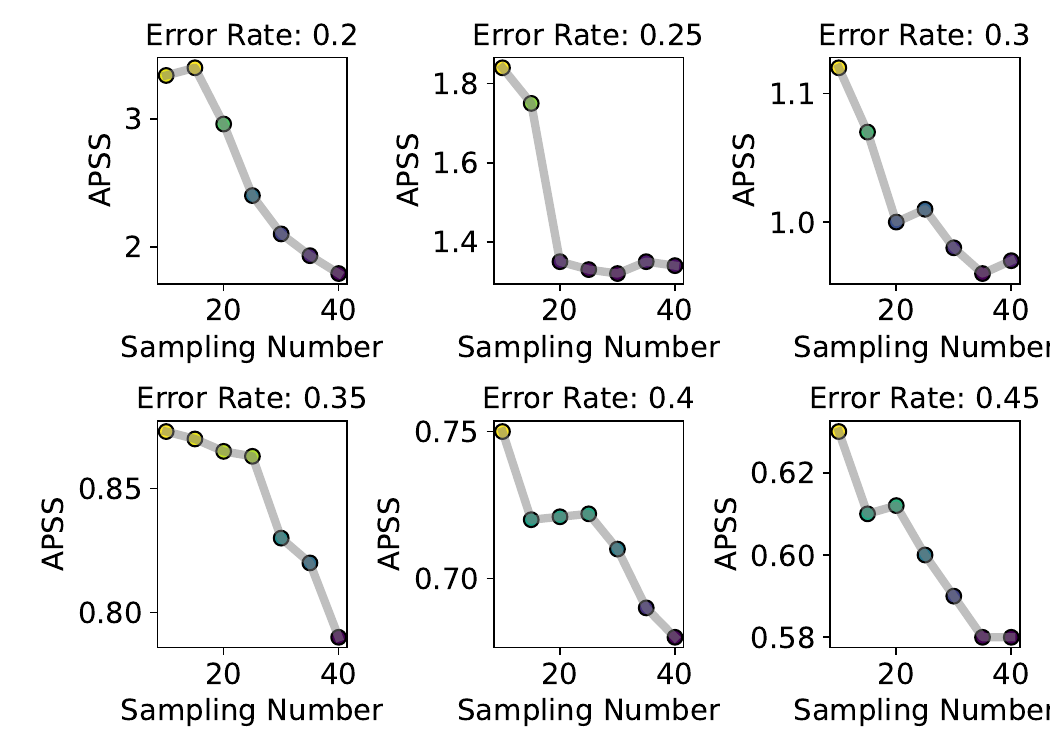}
        \caption{\small All results of the sensitivity analysis to variations in sampling quantity.}
        \label{FIG:sampling-sens-appendix}
    \vspace{-4mm}
\end{figure}

\begin{figure}[h]
        \centering
        \includegraphics[width=\linewidth]{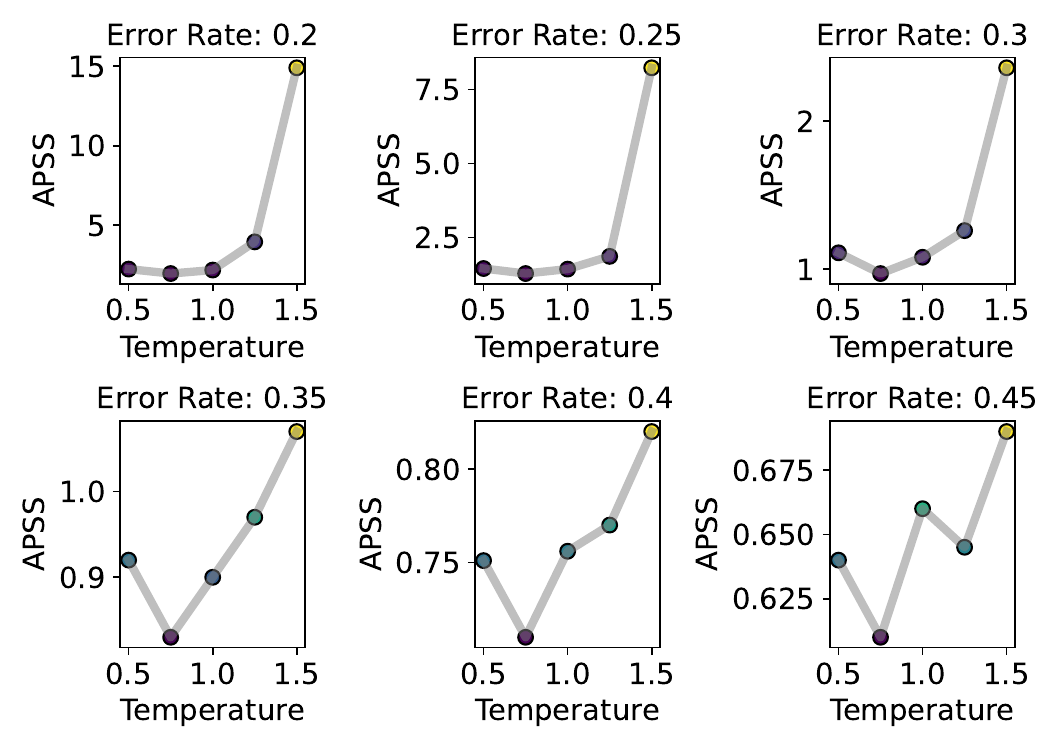}
        \caption{\small All results of the sensitivity analysis to variations in temperature.}
        \label{FIG:temp-sens-appendix}
    \vspace{-2mm}
\end{figure}

More results regarding sampling quantity and temperature sensitivity are included in Figures \ref{FIG:sampling-sens-appendix}-\ref{FIG:temp-sens-appendix} due to the page limit in the main body.

\subsection{Results for WizardLM-v1.2 (13B) and Vicuna-v1.5 (7B)}
To save on computation costs, we use float16 precision (half-precision) for experiments in this section. We use standard sampling with sampling quantity of 30. Results for TriviaQA are shown in Table \ref{results-WizardLM-trivia}, for WebQuestions are shown in Table \ref{results-WizardLM-web}. Results for TriviaQA are shown in Table \ref{results-vc-trivia}, for WebQuestions are shown in Table \ref{results-vc-web}.

\begin{table*}[h]
\caption{Results for TriviaQA using WizardLM-v1.2.} 
\label{results-WizardLM-trivia}
\small
\centering

\begin{subtable}{\textwidth}
\centering
\scalebox{0.85}{%
\begin{tabular}{cccccccccccc} 
\toprule[1pt] 
\multirow{3}{*}{Methods} & \multirow{3}{*}{\shortstack{Logit-Access}} & \multicolumn{9}{c}{Error Rate} \\
& &\multicolumn{3}{c}{0.25} & \multicolumn{3}{c}{0.3} & \multicolumn{3}{c}{0.35} \\
& & ECR & SSC$\uparrow$ & APSS$\downarrow$ & ECR & SSC$\uparrow$ & APSS$\downarrow$ & ECR & SSC$\uparrow$ & APSS$\downarrow$ \\
\midrule 
First-K$_{\text{white}}$ & \ding{51}& 75.1 & 68.7 & 3.19 & 71.0 & 65.8 & 2.56 & 66.4 & 63.3 & 1.84 \\
CLM & \ding{51}  & 75.1 & 63.3 & \textbf{3.01} & 70.1 & 64.9 & 2.20 & 65.0 & 63.3 & 1.43\\
SCP& \ding{51}  & 75.4 & 57.9 & 3.29 & 70.1 & 62.2 & 2.15 & 65.2 & 56.4 & 1.68\\
SAPS & \ding{51}  & 75.1 & \textbf{70.6} & 3.83 & 70.1 & 53.2 & 2.30 & 65.1 & 54.9 & 1.37\\
First-K$_{\text{black}}$& {\ding{55}} & 75.7 & 58.0 & 4.94 & 71.5 & 66.6 & 2.59 & 68.4 & 65.6 & 1.84\\
\textbf{LofreeCP (Ours)} &{\ding{55}} & 75.1 & 68.0 & 4.07 & 70.0 & \textbf{67.7} & \textbf{1.92} & 65.1 & \textbf{70.1} & \textbf{1.27} \\
\end{tabular}
}

\scalebox{0.85}{%
\begin{tabular}{cccccccccccc} 
\toprule[0.65pt] 
\multirow{3}{*}{Methods} & \multirow{3}{*}{\shortstack{Logit-Access}} & \multicolumn{9}{c}{Error Rate} \\
& &\multicolumn{3}{c}{0.4} & \multicolumn{3}{c}{0.45} & \multicolumn{3}{c}{0.5} \\
& & ECR & SSC$\uparrow$ & APSS$\downarrow$ & ECR & SSC$\uparrow$ & APSS$\downarrow$ & ECR & SSC$\uparrow$ & APSS$\downarrow$ \\
\midrule 
First-K$_{\text{white}}$ & \ding{51}& \ding{55} & \ding{55} & \ding{55} & 55.2 & 56.0 & 0.99 & \ding{55} & \ding{55} & \ding{55} \\
CLM & \ding{51}  & 60.1 & 65.3 & 1.25 & 55.1 & 69.1 & 0.92 & 50.1 & 71.3 & 0.81 \\
SCP & \ding{51}  & 60.0 & 65.9 & 1.30 & 55.1 & 67.8 & 1.01 & 50.1 & 70.1 & 0.82 \\
SAPS & \ding{51}  & 60.0 & 47.3 & 1.37 & 55.2 & 53.7 & 1.05 & 50.1 & 60.6 & 0.83 \\
First-K$_{\text{black}}$& {\ding{55}}  & \ding{55} & \ding{55} & \ding{55} & 56.9 & 57.4 & 0.99 & \ding{55} & \ding{55} & \ding{55}\\
\textbf{LofreeCP (Ours)} &{\ding{55}}& 60.2 & \textbf{69.8} & \textbf{0.98} & 55.3 & \textbf{70.4} & \textbf{0.81} & 50.2 & \textbf{72.5} & \textbf{0.69} \\
\bottomrule[1pt] 
\end{tabular}
}
\vspace{-0.3em} 
\end{subtable}
\end{table*}

\begin{table*}[t]
\caption{Results for TriviaQA using Vicuna-v1.5.} 
\label{results-vc-trivia}
\small
\centering

\begin{subtable}{\textwidth}
\centering
\scalebox{0.85}{%
\begin{tabular}{cccccccccccc} 
\toprule[1pt] 
\multirow{3}{*}{Methods} & \multirow{3}{*}{\shortstack{Logit-Access}} & \multicolumn{9}{c}{Error Rate} \\
& &\multicolumn{3}{c}{0.475} & \multicolumn{3}{c}{0.5} & \multicolumn{3}{c}{0.525} \\
& & ECR & SSC$\uparrow$ & APSS$\downarrow$ & ECR & SSC$\uparrow$ & APSS$\downarrow$ & ECR & SSC$\uparrow$ & APSS$\downarrow$ \\
\midrule 
First-K$_{\text{white}}$ & \ding{51}& 53.0 & 42.1 & \textbf{2.23} & 50.4 & 42.4 & 1.63 & \ding{55} & \ding{55} & \ding{55} \\
CLM & \ding{51}  & 52.5 & \textbf{45.1} & 2.60 & 50.1 & 45.5 & 1.39 & 47.5 & 47.7 & 1.21\\
SCP& \ding{51}  & 52.6 & 39.0 & 2.66 & 50.0 & 40.5 & 1.43 & 47.9 & 49.3 & 1.14\\
SAPS & \ding{51}  & 52.7 & 40.1 & 2.30 & 50.3 & \textbf{48.8} & 1.59 & 47.5 & 45.6 & 1.24\\
First-K$_{\text{black}}$& {\ding{55}} & 53.4 & 44.1 & 2.75 & 50.9 & 42.3 & 1.62 & \ding{55} & \ding{55} & \ding{55} \\
\textbf{LofreeCP (Ours)} &{\ding{55}} & 52.5 & 39.3 & 2.27 & 50.0 & 39.1 & \textbf{1.33} & 47.6 & \textbf{50.1} & \textbf{1.12} \\
\end{tabular}
}

\scalebox{0.85}{%
\begin{tabular}{cccccccccccc} 
\toprule[0.65pt] 
\multirow{3}{*}{Methods} & \multirow{3}{*}{\shortstack{Logit-Access}} & \multicolumn{9}{c}{Error Rate} \\
& &\multicolumn{3}{c}{0.4} & \multicolumn{3}{c}{0.45} & \multicolumn{3}{c}{0.5} \\
& & ECR & SSC$\uparrow$ & APSS$\downarrow$ & ECR & SSC$\uparrow$ & APSS$\downarrow$ & ECR & SSC$\uparrow$ & APSS$\downarrow$ \\
\midrule 
First-K$_{\text{white}}$ & \ding{51} & 45.0 & 46.7 & 0.99 & \ding{55}& \ding{55} & \ding{55} & \ding{55} & \ding{55} & \ding{55} \\
CLM & \ding{51}  & 45.2 & 50.7 & 1.01 & 42.5 & 50.6 & 0.85 & 40.1 & 56.2 & 0.83 \\
SCP & \ding{51}  & 45.4 & 52.4 & \textbf{0.96} & 42.6 & 48.6 & 0.85 & 40.5 & 52.0 & 0.76 \\
SAPS & \ding{51}  & 45.0 & 46.2 & 1.04 & 42.6 & 50.8 & 0.84 & 40.1 & 57.9 & 0.75 \\
First-K$_{\text{black}}$& {\ding{55}}  & \ding{55} & \ding{55} & \ding{55} & 44.6 & 46.2 & 0.97 & \ding{55} & \ding{55} & \ding{55}\\
\textbf{LofreeCP (Ours)} &{\ding{55}}& 45.1 & \textbf{55.3} & \textbf{0.96} & 42.7 & \textbf{58.0} & \textbf{0.82} & 40.2 & \textbf{58.5} & \textbf{0.73} \\
\bottomrule[1pt] 
\end{tabular}
}
\vspace{-0.3em} 
\end{subtable}
\end{table*}

\begin{table*}[h]
\caption{Results for WebQuestions using WizardLM-v1.2.}
\label{results-WizardLM-web}
\small
\centering
\begin{subtable}{\textwidth}
\centering
\scalebox{0.85}{%
\begin{tabular}{ccccccccccccccc} 
\toprule[1pt]
\multirow{3}{*}{Methods} & \multirow{3}{*}{\shortstack{Logit-Access}} & \multicolumn{12}{c}{Error rate} \\
& &\multicolumn{3}{c}{0.45} & \multicolumn{3}{c}{0.5} & \multicolumn{3}{c}{0.55} & \multicolumn{3}{c}{0.6}\\
& & ECR & SSC$\uparrow$ & APSS$\downarrow$ & ECR & SSC$\uparrow$ & APSS$\downarrow$ & ECR & SSC$\uparrow$ & APSS$\downarrow$ & ECR & SSC$\uparrow$ & APSS$\downarrow$\\
\midrule 
First-K$_{\text{white}}$ & \ding{51} & 55.5 & 42.5 & 3.40 & 53.0 & 40.6 & 2.70 & 49.1 & 39.0 & 1.91 & \ding{55} & \ding{55} & \ding{55}\\
CLM & \ding{51}  & 55.1 & \textbf{52.3} & 3.02 & 50.2 & 40.1 & 2.01 & 45.2 & 28.6 & 1.58 & 40.4 & 31.2 & 1.19 \\
SCP& \ding{51}  & 55.2 & 45.9 & 3.63 & 50.1 & 40.8 & 2.04 & 45.0 & 37.1 & 1.55 & 40.2 & 47.8 & 1.04\\
SAPS& \ding{51}  & 55.0 & 45.7 & 3.38 & 50.1 & 41.1 & 2.15 & 45.2 & 28.6 & 1.58 & 40.4 & 31.2 & 1.19 \\
First-K$_{\text{black}}$& {\ding{55}}  & 56.7 & 43.6 & 3.40 & 50.9 & 45.0 & 1.91 & \ding{55} & \ding{55} & \ding{55} & 41.4 & 41.1 & 1.00  \\
\textbf{LofreeCP (Ours)} &{\ding{55}}& 55.0 & 45.3 & \textbf{2.87} & 50.0 & \textbf{46.5} & \textbf{1.88} & 45.1 & \textbf{49.9} & \textbf{1.18} & 40.1 & \textbf{51.7} & \textbf{0.82} \\
\bottomrule[1pt] 
\end{tabular}
}
\end{subtable}
\vspace{-0.3em}
\end{table*}

\begin{table*}[h]
\caption{Results for WebQuestions using Vicuna-v1.5.}
\label{results-vc-web}
\small
\centering
\begin{subtable}{\textwidth}
\centering
\scalebox{0.85}{%
\begin{tabular}{ccccccccccccccc} 
\toprule[1pt]
\multirow{3}{*}{Methods} & \multirow{3}{*}{\shortstack{Logit-Access}} & \multicolumn{12}{c}{Error rate} \\
& &\multicolumn{3}{c}{0.575} & \multicolumn{3}{c}{0.6} & \multicolumn{3}{c}{0.625} & \multicolumn{3}{c}{0.65}\\
& & ECR & SSC$\uparrow$ & APSS$\downarrow$ & ECR & SSC$\uparrow$ & APSS$\downarrow$ & ECR & SSC$\uparrow$ & APSS$\downarrow$ & ECR & SSC$\uparrow$ & APSS$\downarrow$\\
\midrule 
First-K$_{\text{white}}$ & \ding{51} & 43.2 & 23.8 & 1.99 & 41.7 & 26.9 & 1.57 & \ding{55} & \ding{55} & \ding{55} & 36.6 & 36.6 & 1.00\\
CLM & \ding{51}  & 42.5 & 32.3 & 1.88 & 40.1 & 36.2 & 1.32 & 37.6 & 38.2 & 1.08 & 35.0 & 41.8 & 0.83 \\
SCP& \ding{51}  & 42.6 & 31.1 & 1.91 & 40.1 & 34.4 & 1.28 & 38.2 & 37.3 & 1.06 & 35.2 & \textbf{43.7} & 0.87\\
SAPS& \ding{51}  & 42.5 & 32.3 & 1.88 & 40.1 & 36.2 & 1.32 & 37.6 & 38.2 & 1.08 & 35.0 & 41.8 & 0.83 \\
First-K$_{\text{black}}$& {\ding{55}}  & 43.7 & 25.9 & 2.01 & 40.9 & 25.5 & 1.57 & \ding{55} & \ding{55} & \ding{55} & 36.8 & 36.8 & 1.00  \\
\textbf{LofreeCP (Ours)} &{\ding{55}}& 42.5 & \textbf{32.4} & \textbf{1.73} & 40.1 & \textbf{36.7} & \textbf{1.22} & 37.5 & \textbf{39.6} & \textbf{0.97} & 35.0 & 39.3 & \textbf{0.81} \\
\bottomrule[1pt] 
\end{tabular}
}
\end{subtable}
\vspace{-0.3em}
\end{table*}

Results for WizardLM-v1.2 (13B) and Vicuna-v1.5 (7B) consistently align with the main body results, demonstrating that the LofreeCP method mostly outperforms the baselines.

\end{document}